\documentclass{article}
\usepackage{ijcai26}

\usepackage{times}
\usepackage{soul}
\usepackage{url}
\usepackage[hidelinks]{hyperref}
\usepackage[utf8]{inputenc}
\usepackage[small]{caption}
\usepackage{graphicx}
\usepackage{amsmath}
\usepackage{amsthm}
\usepackage{booktabs}
\usepackage{algorithm}
\usepackage{algorithmic}
\usepackage[switch]{lineno}
\usepackage{comment}
\usepackage{subcaption}

\usepackage{microtype}

\usepackage{stfloats}

\usepackage{tikz}
\usetikzlibrary{matrix}
\usepackage{amssymb}
\usetikzlibrary{positioning}
\usetikzlibrary{positioning, shapes.geometric, arrows.meta, fit, calc}
\usepackage{multirow}
\usepackage{adjustbox}
\usepackage[normalem]{ulem}

\newlength{\imgwidth}

\newlength{\gridimg}

\usepackage{xcolor}

\title{FlowID: Enhancing Forensic Identification with Latent Flow-Matching Models}

\author{
Jules Ripoll$^1$
\and
David Bertoin$^1$\and
Alasdair Newson$^{2}$\and
Charles Dossal$^1$\And
Jose Pablo Baraybar$^3$\\
\affiliations
$^1$INSA Toulouse\\
$^2$Sorbonne Université\\
$^3$International Committee of the Red Cross
\emails
\{jripoll, bertoin, dossal\}@insa-toulouse.fr,
anewson@isir.upmc.fr,
jpbaraybar@icrc.org
}

\begin{document}
\looseness=-1
%\raggedbottom
\maketitle

\begin{abstract}

Every day, many people die under violent circumstances, whether from crimes, war, migration, or climate disasters. 
Medico-legal and law enforcement institutions document many portraits of the deceased for evidence, but cannot immediately carry out identification on them. 
While traditional image editing tools can process these photos for public release, the workflow is lengthy and produces suboptimal results. 
In this work, we leverage advances in image generation models, which can now produce photorealistic human portraits, to introduce FlowID, an identity-preserving facial reconstruction method. Our approach combines single-image fine-tuning, which adapts the generative model to out-of-distribution injured faces, with attention-based masking that localizes edits to damaged regions while preserving identity-critical features. 
Together, these components enable the removal of artifacts from violent death while retaining sufficient identity information to support identification. 
To evaluate our method, we introduce InjuredFaces, a novel benchmark for identity-preserving facial reconstruction under severe facial damage. 
Beyond serving as an evaluation tool for this work, InjuredFaces provides a standardized resource for the community to study and compare methods addressing facial reconstruction in extreme conditions.
Experimental results show that FlowID outperforms state-of-the-art open-source methods while maintaining low memory requirements, making it suitable for local deployment without compromising data privacy.

\end{abstract}
\noindent\fbox{\parbox{\dimexpr\columnwidth-2\fboxsep-2\fboxrule}{%
\textbf{Content warning:} This paper contains images of facial injuries that readers may find unpleasant.}}
\section{Introduction}

\newlength{\teaserimg}
\begin{figure}[t]
\centering
\setlength{\tabcolsep}{3pt}
\setlength{\teaserimg}{0.22\columnwidth}
\begin{tabular}{cccc}
% Column 1
\includegraphics[width=\teaserimg]{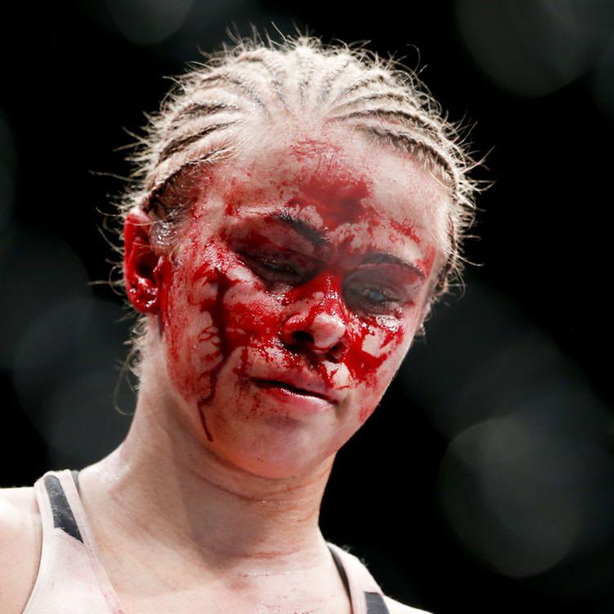} &
\includegraphics[width=\teaserimg]{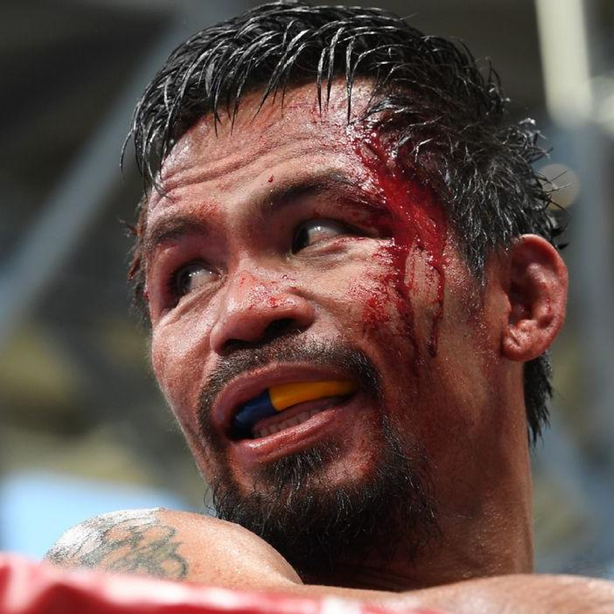} &
\includegraphics[width=\teaserimg]{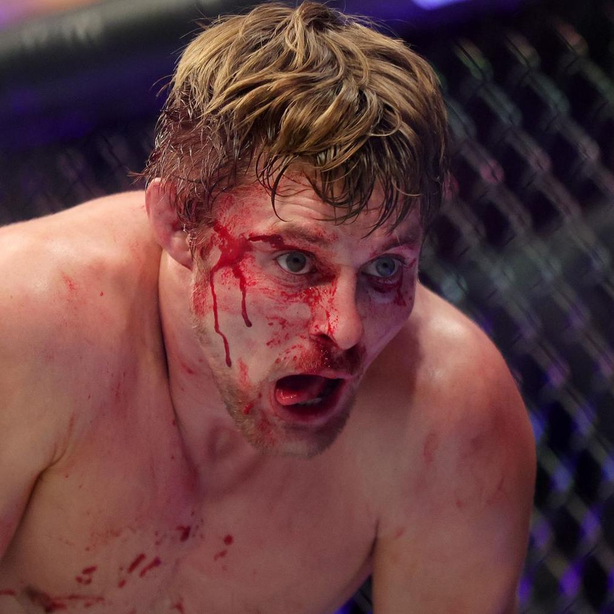} &
\includegraphics[width=\teaserimg]{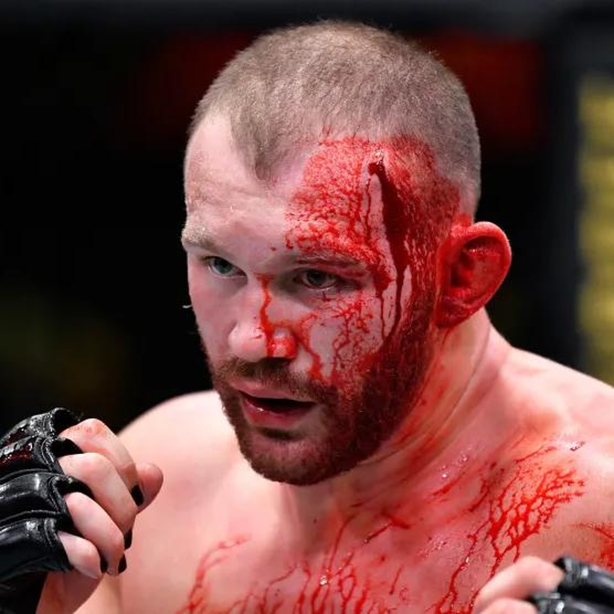} \\[0pt]
\includegraphics[width=\teaserimg]{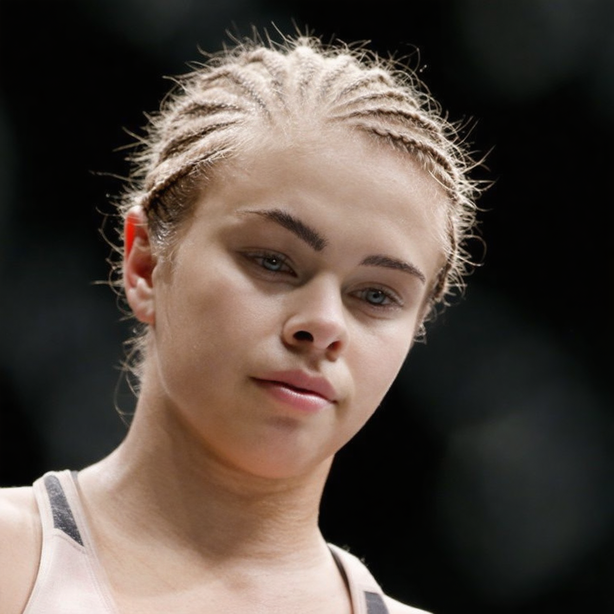} &
\includegraphics[width=\teaserimg]{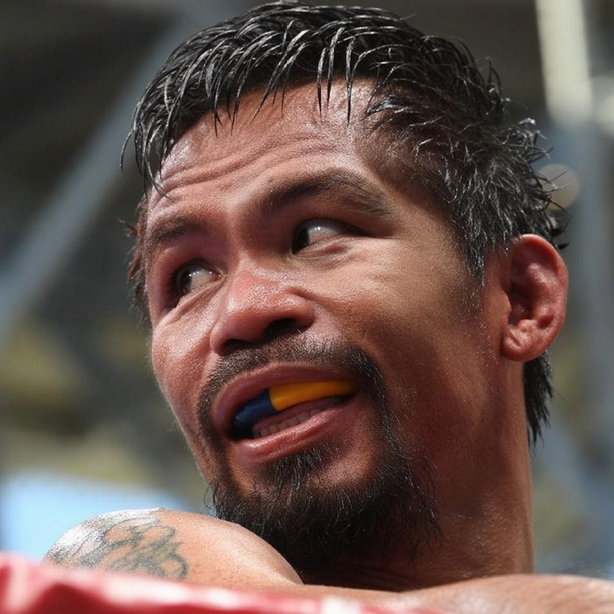} &
\includegraphics[width=\teaserimg]{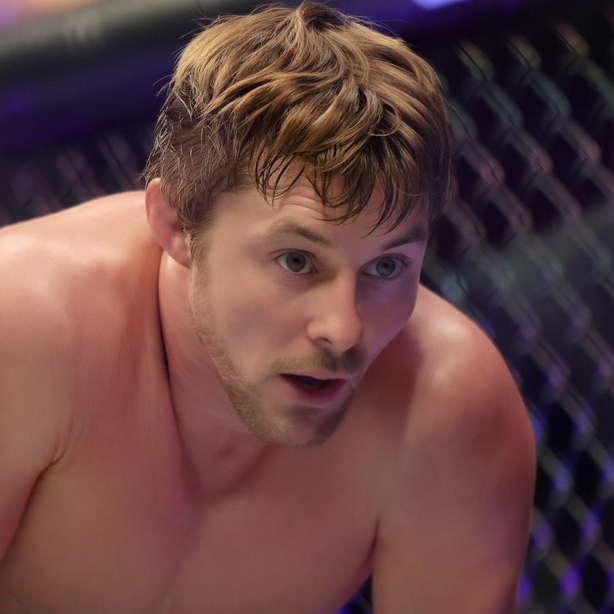} &
\includegraphics[width=\teaserimg]{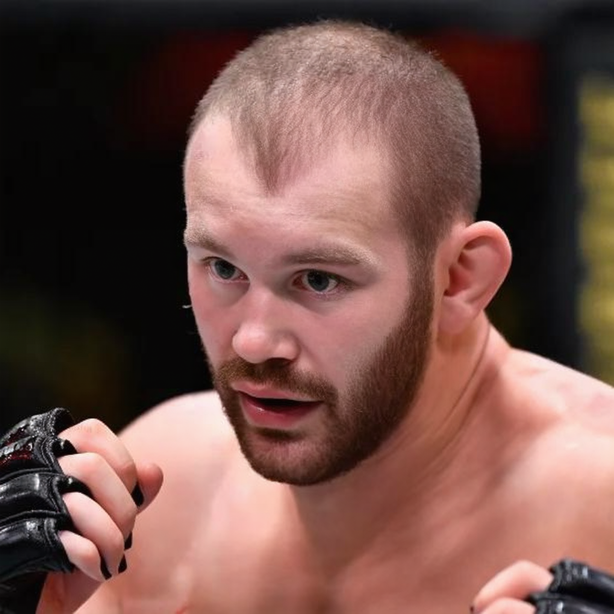}
\end{tabular}
\caption{Samples from InjuredFaces, our facial reconstruction benchmark, edited with FlowID.
Top: original images. Bottom: edited results.}
\label{fig:teaser}
\end{figure}

Visual recognition by next-of-kin remains the first and most critical step in identifying human corpses for medico-legal institutions worldwide.
In practice, photographs of the deceased are shown to potential relatives, who assess whether they recognize the individual.
In cases of violent deaths, said photographs can be extremely graphic and cause severe psychological distress to the person seeing the picture, and may be unnecessary if the recognition ultimately fails.
To circumvent this issue, and enable broader dissemination, traditional image editing tools are used to remove visible death-related artifacts.
Unfortunately, this method is labor-intensive and only produces satisfying results for minor injuries, failing for severe or structural facial trauma.
This limitation is particularly problematic, given the urgent need for efficiency gains in medico-legal workflows.
In Mexico alone, more than 72,000 bodies remain unidentified alongside 130,000 reported missing persons.

Given the scale of data needed to be processed for identification, semi-automatic tools that accelerate and make the identification process easier would provide substantial societal benefit.
This motivates the following question : \emph{``Can generative model-based image editing be effectively applied to facial reconstruction to facilitate identification by next-of-kin?''} 
This question was inspired by efforts by the International Committee of the Red Cross (ICRC) to address the migration crisis across the Mediterranean. Initiatives
like ``Trace the Face''\footnote{\url{https://tracetheface.familylinks.icrc.org/?lang=en}} — a virtual photo gallery enabling people to search for loved ones lost during migration—have reconnected 309 families out of 7,677 published photographs, but remain primarily designed for the living. 
The challenge of identifying the deceased is far more complex: bodies are often recovered without identification documents or any information concerning their identity or whereabouts, and photographs may need to be shared with communities of origin or the diaspora to stimulate recognition. This problem extends beyond migration to other forensic settings where facial trauma impedes visual recognition.

Addressing this challenge has only recently become feasible due to rapid advances in image generation.
Modern generative models achieve high visual fidelity and fine-grained semantic control, making them suitable candidates for controlled facial reconstruction.
Recent progress is largely driven by diffusion \cite{sohl2015deep,ho2020denoising,song2021scorebased} and flow-matching models \cite{liu2022flow,lipman2023flow}, enhanced by transformer-based architectures \cite{Peebles_2023_ICCV,esser2024scaling,flux2024}.
Despite these advances, deploying generative image editing in a forensic context poses substantial challenges.
First, reconstructed faces must remain as faithful as possible to the original identity except for the death-related artifacts that are intentionally removed.
This tension between preserving fidelity and enabling meaningful edits is commonly referred to as the editability-reconstruction trade-off and lies at the core of image editing.
Second, processing of such sensitive data must be performed locally, using relatively common hardware, to satisfy privacy and operational constraints.
This requirement is non-trivial as editing performance typically scales with model size and computational cost.

In this work, we address both of these challenges and demonstrate that generative image editing can be integrated into forensic identification procedures, to support recognition by next-of-kin.
We introduce InjuredFaces, the first benchmark for evaluating generative facial reconstruction under severe damage, and propose FlowID, an algorithm that outperforms other open-source methods while remaining lightweight enough for deployment on consumer hardware.
The deployment of our algorithm has started in various locations in South America, enabling the identification of several persons.
\section{Related Work}
\begin{comment}
Research on image editing with diffusion and flow matching models has developed along two complementary directions.
Training-free methods achieve editing by manipulating the generation dynamics of pre-trained models, requiring no additional training.
On the other hand, training-based methods expand the model’s capabilities by optimizing new parameters, through fine-tuning or the introduction of adapter networks.  
\end{comment}
\paragraph{Training-free image editing.} Training-free methods mostly rely on perturbing the source image via a forward noising process to project it to a more editable state, and then applying a reverse denoising process to steer the image towards a target edit.
SDEdit \cite{meng2022sdedit} was the first to demonstrate this idea to produce edits where coarse guiding sketches were transformed into realistic samples.
However, large edits require sending the source image to strong noise levels to be faithfully applied, which significantly diminishes the fidelity to the input.
A major step forward came with the introduction of deterministic solvers, such as in \cite{song2021scorebased} and DDIM \cite{SongME21}.
Beyond accelerating sampling, these solvers allow for image inversion, that is, mapping the image back to its latent representation by reversing the denoising process.
When the image is well aligned with the model distribution, inversion yields a likely generation trajectory that can be reused for editing, and this principle has since become the foundation of many training-free approaches.
Building on this, several work refine inversion for editing. 
DiffEdit \cite{couairon2022diffeditdiffusionbasedsemanticimage} combines automatic mask generation with DDIM inversion, ensuring perfect reconstruction in unedited regions.
Other methods exploit reference cross-attention maps to better control the layout during editing : \cite{parmar2023zero} aligns attention maps through gradient steps, while Prompt-to-Prompt \cite{hertz2022prompt} directly substitutes maps from the reference prompt.
Null-Text Inversion \cite{mokady2023null} extends these techniques to real images by optimizing null embeddings during DDIM inversion.
More recent approaches further improve fidelity: DDPM inversion \cite{huberman2024edit} achieves exact reconstruction by storing noise maps, while LEdits++ \cite{brack2024ledits++} restricts edits using masks derived from cross-attention.
TightInversion \cite{kadosh2025tight} uses network adapters to condition inversion with an image to improve reconstruction quality while retaining editability.
RF-Solver \cite{wang2024taming} increases inversion accuracy by employing higher-order ODE solvers, albeit at higher computational cost.
Despite their effectiveness, training-free methods often require careful hyperparameter tuning and remain sensitive to inversion quality, which can limit their robustness and reliability in practice.  
FlowID addresses the issue of inversion quality with a light single-image fine-tuning, achieving a better balance between editability and preservation of the original image, while retaining a relatively low hyperparameter count.
\paragraph{Training-based image editing.} This framework takes a different approach by directly learning the parameters of the model dedicated to image editing, which generally translate to better editing performance.
ICEdit \cite{zhang2025context} leverages the inherent capability of modern text-to-image DiTs \cite{esser2024scaling}\nocite{flux2024} to generate coherent panels and further improves the editing performance by learning LoRA \cite{hu2022lora} parameters to better follow editing instructions.
Another research direction focuses on instruction-based image editing.
InstructPix2Pix \cite{brooks2023instructpix2pix}, EmuEdit \cite{sheynin2024emu}, and UltraEdit \cite{zhao2024ultraedit} curate large datasets of source images, target images, and edit instructions, allowing diffusion models to incorporate editing prompts directly during training.
Flux Kontext \cite{batifol2025flux} extends this strategy with a rectified flow transformer and concatenates text tokens from the instruction prompt with image tokens from the source image, thus allowing mutual interaction during the joint-attention operation.
Finally, some methods adapt the model at edit time to better align with a specific source image.
Unitune \cite{10.1145/3592451} fine-tunes the noise predictor on the source image and accompanying text description, and then applies SDEdit for editing.
Imagic \cite{kawar2023imagic} optimizes both the prompt embedding and the model weights, enabling smooth interpolation between the source and target prompts.
Training-based methods often achieve stronger semantic control and sometimes higher edit fidelity than training-free methods, but require additional training, heavier computation and rely on efficient data collection or carefully designed strategies to construct pairs of source images, edited images, and editing instructions, which further limits their scalability.
Our method overcomes this by fine-tuning on a single image, ensuring a well-integrated edit.

\paragraph{Deep learning in forensics.} Deep learning has also been applied to forensic workflows. Notably, \cite{9306878,majumdar2019subclass} adapt facial recognition networks to injured faces, enabling embedding-based matching against reference databases. While these methods operate at the embedding level for algorithmic matching, FlowID instead modifies the image itself, producing a reconstruction suitable for recognition by next-of-kin.

\section{Method}
\subsection{Background}
\paragraph{Latent Flow-Matching Models.}Modern text-to-image flow matching models generate samples from a data distribution $p_0$ by iteratively denoising a Gaussian noise sample $z_t$.
This process is described by the \textit{probability flow ODE}:

\begin{equation}
\frac{dz_t}{dt}=v(z_t,t),
\label{eq:probabilityflowode}
\end{equation}

where the right-hand term, called the \emph{velocity}, can be learned via the conditional flow matching objective:

\begin{equation}
\mathcal{L}_{CFM} = \mathbb{E}_{t,z_0,\epsilon}\lVert v_{\theta}(z_t,t)-(\dot{\alpha_t}z_0+\dot{\sigma_t}\epsilon)\rVert_2^2
\label{eq:losscfm}
\end{equation}
where $v_\theta$ denotes the learned velocity, $\epsilon \sim \mathcal{N}(0,I_d)$, and $z_t=\alpha_tz_0+\sigma_t\epsilon$ is an interpolation between $\epsilon$ and $z_0$.
For the linear interpolant $\alpha_t = 1 - t$ and $\sigma_t = t$, which is the most commonly used, the target velocity simplifies to $\epsilon - z_0$. The aforementioned processes are carried out in the latent space of a VAE with encoder $\mathcal{E}$ and decoder $\mathcal{D}$, such that for a given image $I$, $z_0 = \mathcal{E}(I)$ and $I = \mathcal{D}(z_0)$.
Additionally, the velocity prediction $v_\theta(z_t,t)$ can be influenced by supplementary conditions such as a text prompt, which we denote as $P$.

\paragraph{Image Inversion.}  A common strategy for image editing with text-conditioned flow matching models relies on inversion. In this setting, one can solve~\eqref{eq:probabilityflowode} forward in time, starting from $t=0$, with the source latent $z_0^{src}$ and the source description $P^{src}$, up to a timestep $t_s$, referred to as the \emph{strength}.
Then, editing can be achieved by solving~\eqref{eq:probabilityflowode} again, in reverse time, starting from $z_{t_s}$ and using the target textual conditioning $P^{tgt}$ to obtain the edited latent $z_0^{tgt}$.

\begin{figure*}[t]
    \centering
    \setlength{\tabcolsep}{2pt}
    \setlength{\imgwidth}{0.12\textwidth}
    \begin{tabular}{ccccccc}
        & $\mathrm{MSE=0.09}$ & $0.06$ & $0.06$ & $0.08$ & $0.05$ & $0.02$ \\
        \includegraphics[width=\imgwidth]{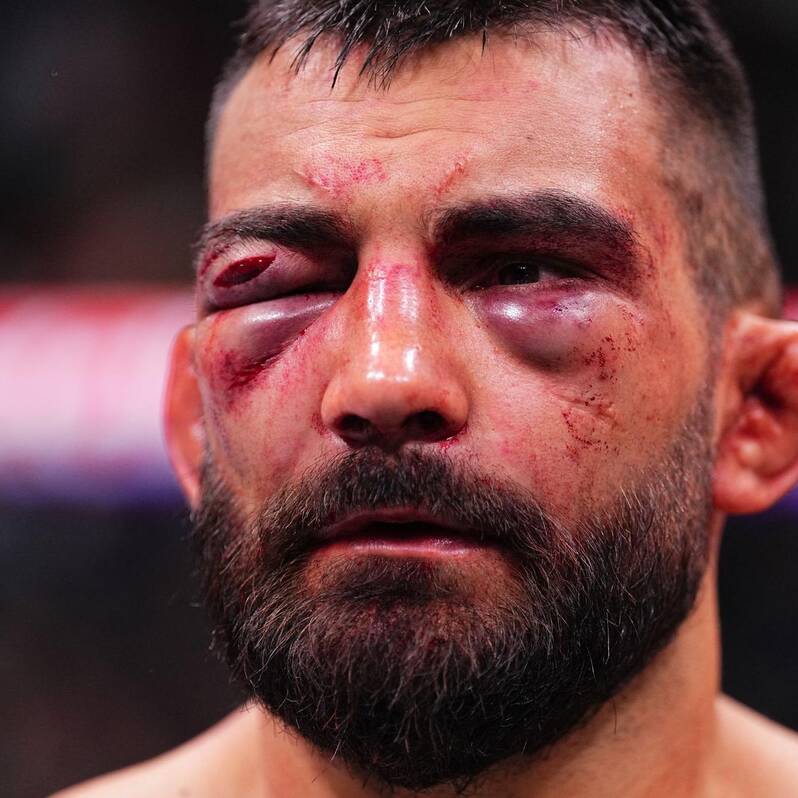} &
        \includegraphics[width=\imgwidth]{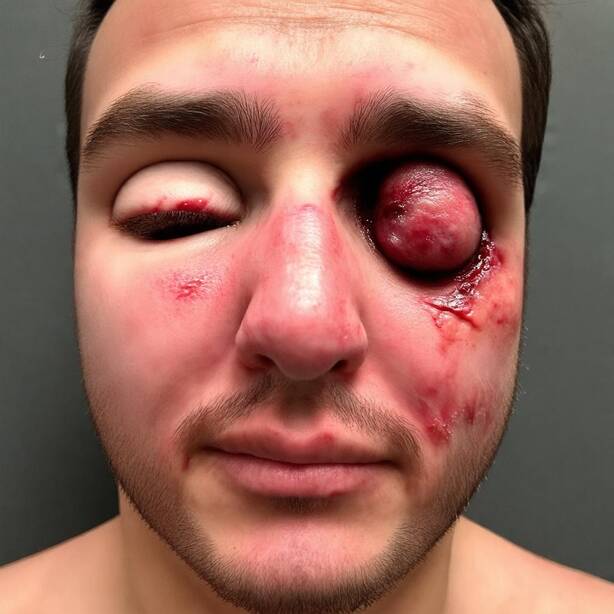} &
        \includegraphics[width=\imgwidth]{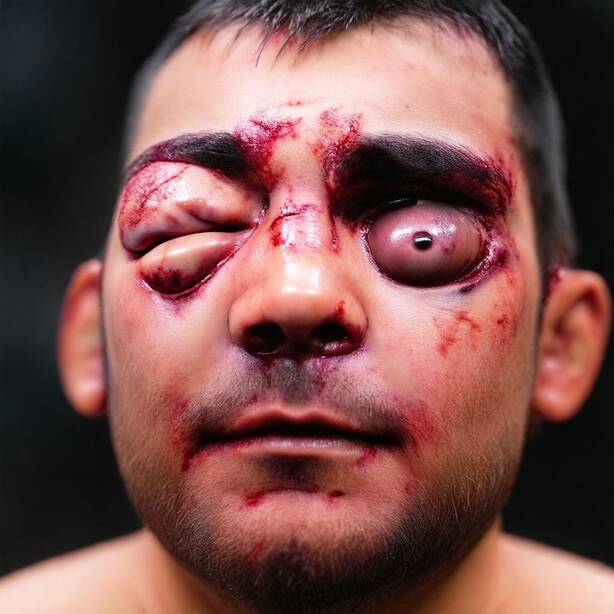} &
        \includegraphics[width=\imgwidth]{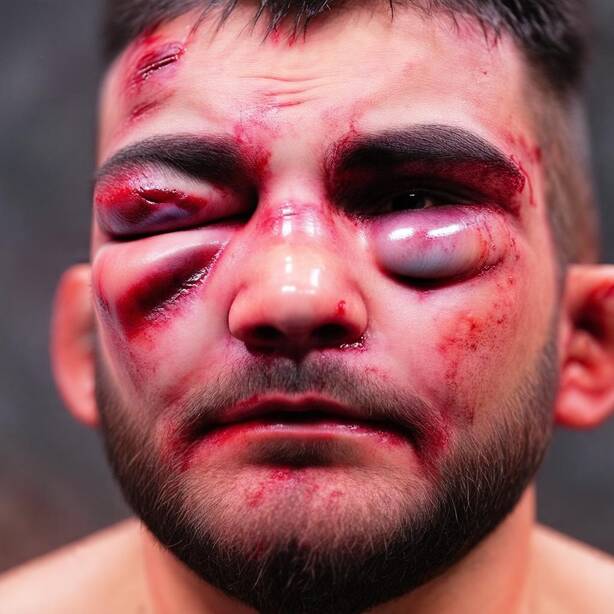} &
        \includegraphics[width=\imgwidth]{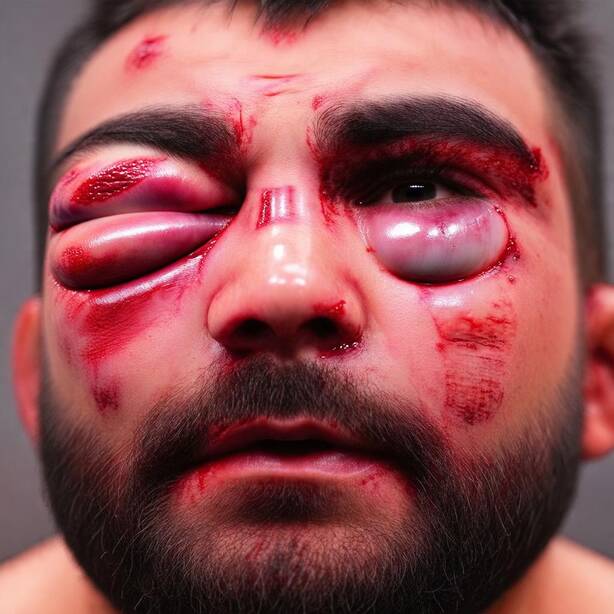} &
        \includegraphics[width=\imgwidth]{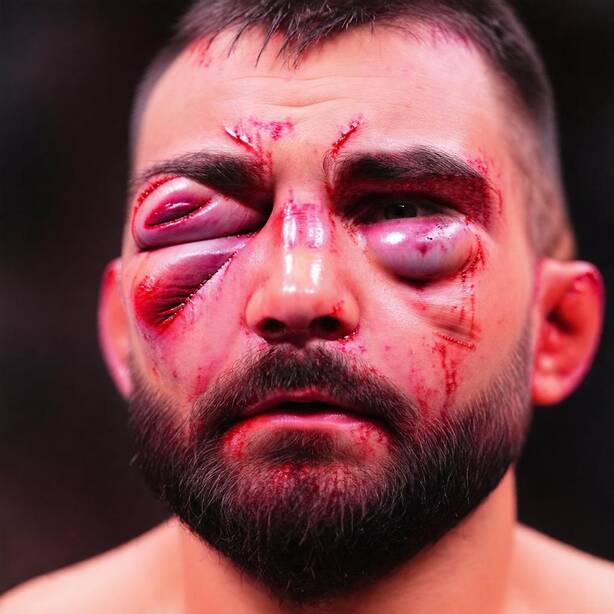} &
        \includegraphics[width=\imgwidth]{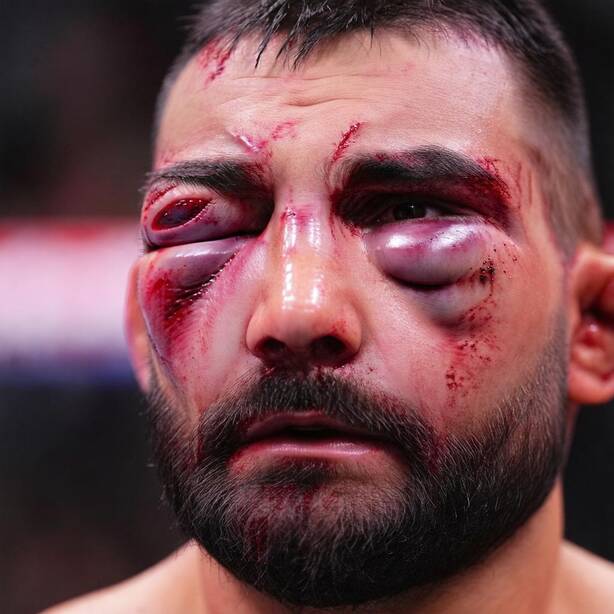} \\
        $\mathrm{Input}$ & $\mathrm{N=0}$ & $20$ & $40$  & $60$  & $80$  & $100$ 
    \end{tabular}
    \caption{Single-image fine-tuning improves inversion performance.}
    \label{fig:ft_improves_inversion}
\end{figure*}

\subsection{Single-Image Fine-Tuning}
\label{sec:fine-tuning}

Editing large regions using inversion-based methods requires integrating Equation \eqref{eq:probabilityflowode} up to a large $t_s$.
As previously described, this poses the problem of error accumulation when using numerical ODE solvers for the inversion.
If the inversion is imperfect, crucial identity information is lost and cannot be recovered during reconstruction.
We argue that part of the error induced in the inversion stems from the fact that the image to edit is not a sample generated by the model and the semantic alignment between the inversion prompt and the image is suboptimal.
Our method begins with single-image fine-tuning on the user-supplied image and prompt, using the pretrained Stable Diffusion 3 \cite{esser2024scaling} flow-matching model. This procedure adjusts the model such that the provided image becomes a more likely sample under the conditioning prompt $P^{src}$. Technically, this amounts to applying the usual flow-matching objective (Eq.~\ref{eq:losscfm}) to just one image–prompt pair.

Figure~\ref{fig:ft_improves_inversion} shows the effect of single-image fine-tuning on inversion quality, measured by the MSE between the input and reconstructed images.
We perform full inversion on the original image, stopping slightly before reaching pure noise (2 steps), in order to retain structural information.
% We perform full inversion, on the original image, except for the final two integration steps, which are omitted to better preserve the reference inversion trajectory.
As the number of fine-tuning steps $N$ increases, reconstruction error decreases steadily.
Fine-tuning progressively recovers global structure, with head pose and identity-related attributes such as facial hair emerging early in the process.
\subsection{Localized Editing with Inferred Masks}

Recognizing a person from an image is based on visible morphological characteristics, \emph{e.g} the shape of the nose, mouth, jaw, or the distance between the eyes. However, recognition can also depend on secondary cues such as tattoos or jewelry. In practice, these fine-grained details are often not faithfully preserved by the inversion and generation process, and may be inadvertently altered during reconstruction and thus hinder recognition.
To address these limitations, our approach incorporates \textit{localized editing}, which explicitly preserves regions of the image that should remain unchanged.

Deciding which areas to preserve and which to modify is a non-trivial challenge, particularly when relying only on textual prompts. 
To overcome this, we exploit internal signals from the model itself.
FlowID leverages the attention maps of the generative model to construct the mask $M$.

\begin{figure}[t]
\centering
\setlength{\tabcolsep}{2pt}
\setlength{\fboxsep}{0pt}
\setlength{\fboxrule}{0.3pt}
\resizebox{0.9\columnwidth}{!}{%
\begin{tabular}{@{}ccc@{}}
% Top row - masks
& 
\fbox{\includegraphics[width=2cm]{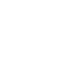}} &
\fbox{\includegraphics[width=2cm]{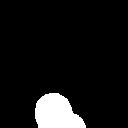}} \\[2pt]
% Middle row - main images
\begin{subfigure}[b]{2.5cm}
    \centering
    \includegraphics[width=\textwidth]{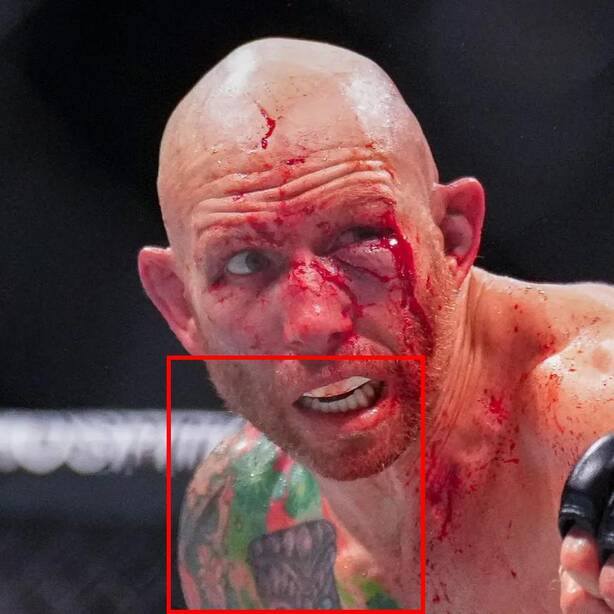}
\end{subfigure} &
\begin{subfigure}[b]{2.5cm}
    \centering
    \includegraphics[width=\textwidth]{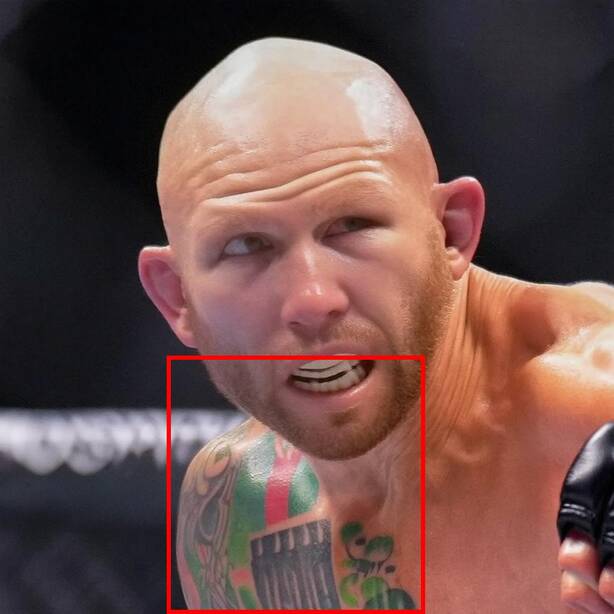}
\end{subfigure} &
\begin{subfigure}[b]{2.5cm}
    \centering
    \includegraphics[width=\textwidth]{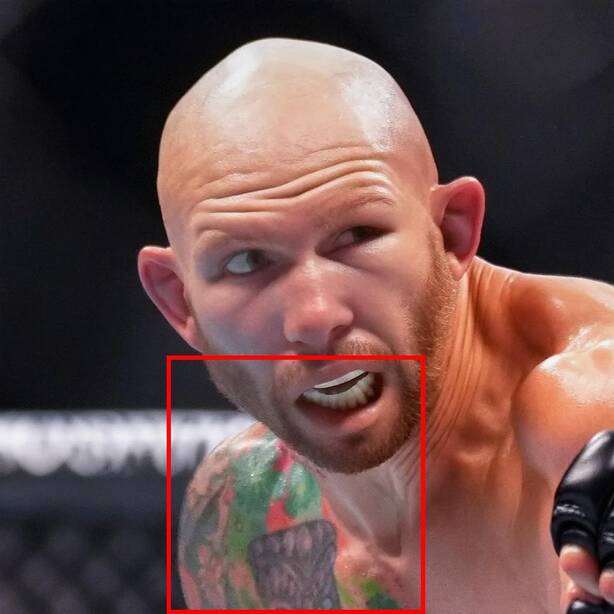}
\end{subfigure} \\[2pt]
% Bottom row - cropped images with captions
\begin{subfigure}[b]{2.5cm}
    \centering
    \includegraphics[width=\textwidth]{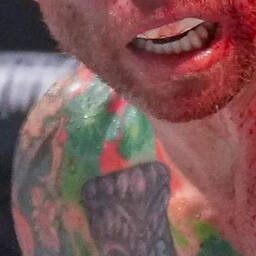}
    \caption*{\scriptsize Input}
\end{subfigure} &
\begin{subfigure}[b]{2.5cm}
    \centering
    \includegraphics[width=\textwidth]{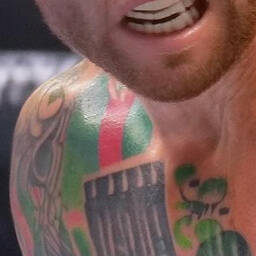}
    \caption*{\scriptsize w/o Masking}
\end{subfigure} &
\begin{subfigure}[b]{2.5cm}
    \centering
    \includegraphics[width=\textwidth]{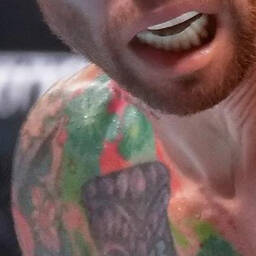}
    \caption*{\scriptsize w/ Masking}
\end{subfigure}
\end{tabular}
}%
\caption{Impact of our masking component for details preservation. Without masking, the person's tattoo is altered \emph{(middle)} When using our masking component, we manage to infer the location of the tattoo and explicitly preserve it \emph{(right).}}
\label{fig:mask-based-preservation}
\end{figure}
Prior work has shown that the intermediate layers of transformer blocks capture rich semantic information \cite{Tumanyan_2023_CVPR,luo2023diffusion,epstein2023diffusion,helbling2025conceptattention}, which can be repurposed to guide image editing \cite{hertz2022prompt,parmar2023zero,epstein2023diffusion,brack2024ledits++}.

For every transformer block in the DiT, consider a sequence of image tokens $z \in \mathbb{R}^{h \times w \times d}$ and of text tokens $c \in \mathbb{R}^{l \times d}$.
Inside this text tokens sequence is a tokens subsequence corresponding to the concept of interest that we wish to edit, denoted as $c^* \in \mathbb{R}^{k \times d}$, with $k \le l$.

Then, for a given attention head, we extract the query $q_z$ for all the image tokens, as well as the key $k_{c^{\ast}}$ for $c^\ast$:
\begin{equation}
q_z=Q_zz, \:  k_{c^*}=K_cc^*
\end{equation}
We compute the image concept interaction map $\alpha \in \mathbb{R}^{h \times w \times k}$ as $\alpha = \mathrm{softmax}(q_z k_{c^*}^T).$

By averaging over concept tokens and reshaping the obtained vector into a matrix, we obtain a cross-attention map $m_{a,b} \in \mathbb{R}^{h \times w}$, for head $a$ and block $b$, indicating where the concept is present in $z$.
We then average over all attention heads and all transformer blocks, which yields a more salient map $\bar{m}$.
Finally, we binarize $\bar{m}$ with a K-means algorithm with 2 barycenters, effectively capturing the binary nature of $\bar{m}$, to produce our final mask $M \in \{0,1\}^{h \times w}$.

Recomputing the mask at every generation step proves ineffective, as the signal tends to vanish toward the end of the backward process. Instead, we compute it once at a specific step $t_M$, chosen such that $t_M < t_s$, ensuring that the object to be edited is well localized set at the time of mask estimation. For concept removal tasks, this condition is naturally satisfied since the mask can be obtained during inversion. For concept addition, however, we proceed differently: the backward process is first run from $t_s$ down to $t_M$ without masking, the mask is then computed at $t_M$, and finally the process is backtracked to $t_s$ before resuming the actual editing.

\cite{couairon2022diffeditdiffusionbasedsemanticimage} noticed that it is beneficial to have a mask that overshoots the area to be edited.
We provide such property to FlowID by applying a max pooling operation to make it bigger.
\begin{comment}
The pasting operation can sometimes leave artifacts in the final image, where the limit of the mask can be spotted. 
We counteract this by smoothing the mask, using a Gaussian blur operation, to make the edited and preserved parts merge more smoothly.
\end{comment}

Figure~\ref{fig:mask-based-preservation} illustrates the effect of our masking strategy when removing blood while preserving a body tattoo.
Without masking, the generative process alters fine tattoo details, reducing recognition reliability.
In contrast, explicit masking prevents unintended modifications by enforcing exact reconstruction within the tattoo region.

To summarize, let us consider the following reference inversion trajectory under $v_\theta$ and a given encoded prompt $c^{src}$ : $\{z_0^{ref},z_{t_1}^{ref},\dots,z_{t_s}^{ref}\}$.
Assuming we have a binary mask $M$ that separates areas that should be edited from areas that must remain untouched, we paste reference parts on the current ODE solver iterate $z_{t-1}'$: $z_{t-1}=M*z_{t-1}'+(1-M)*z_{t-1}^{ref}$ in a similar fashion to \cite{couairon2022diffeditdiffusionbasedsemanticimage}. A pseudocode for the FlowID method can be seen in Algorithm~\ref{algo:FlowID}.

\begin{algorithm}[tb]
    \caption{FlowID algorithm}
    \label{algo:FlowID}
    \textbf{Input}: $v_\theta$, $z_0^{src}$, $c^{src}$, $c^{tgt}$, $c^*$, $t_s$, $\{\sigma_t\}_t$\\
    \textbf{Output}: $z_0^{tgt}$
    \begin{algorithmic}[1]
        \STATE $v_\theta \gets \textsc{FineTune}(v_\theta, z_0^{src}, c^{src})$
        \STATE $z_t \gets z_0^{src}$
        \STATE $z_0^{ref} \gets z_0^{src}$
        \FOR{$t = 0, \ldots, t_{s-1}$}
            \STATE $z_{t+1} \gets z_t+(\sigma_{t+1} - \sigma_t) v_\theta(z_t, t, c^{src})$
            \STATE $z_{t+1}^{ref} \gets z_{t+1}$
        \ENDFOR
        \STATE $M \gets \textsc{GetMask}(\{z_t^{ref}\}_t, c^{src}, c^{tgt}, c^*)$
        \FOR{$t = t_s, \ldots, t_1$}
            \STATE $z_{t-1}' \gets z_t+(\sigma_{t-1} - \sigma_t) v_\theta(z_t, t, c^{tgt})$
            \STATE $z_{t-1} \gets M \cdot z_{t-1}' + (1-M) \cdot z_{t-1}^{ref}$
        \ENDFOR
        \STATE $z_0^{tgt} \gets z_t$
        \STATE \textbf{return} $z_0^{tgt}$
    \end{algorithmic}
\end{algorithm}
\section{Experiments}

In this section, we present a comprehensive experimental evaluation of our approach.
We first introduce InjuredFaces, a new benchmark for facial reconstruction under severe injuries, explicitly designed to evaluate identity preservation.
\begin{comment}
    InjuredFaces is a core contribution of this work and provides the first standardized evaluation protocol for generative facial reconstruction in forensic contexts.
\end{comment}
We then evaluate FlowID using Stable Diffusion~3 and compare it against recent state-of-the-art editing methods, including UltraEdit, SDEdit, ICEdit, Kontext, and RF-Solver.
To assess generalization beyond injury reconstruction, we further evaluate our method on classical and sequential facial editing tasks using a subset of the FFHQ dataset \cite{karras2019style}.
Finally, we present an ablation study to quantify the contribution of each component of our approach.
Additional hyperparameters and implementation details are provided in Appendix~\ref{experiment_details}.

\subsection{The InjuredFaces Benchmark}
\label{sec:injuredfaces}
Evaluating facial reconstruction quality is essential for forensic identification, yet reproducible evaluation is hindered by the impossibility of sharing real forensic images due to privacy and ethical constraints.
To overcome this barrier, we introduce InjuredFaces, a new benchmark for generative facial reconstruction under severe injuries, built from publicly available images of professional athletes.
InjuredFaces is a core contribution of this work and is intended as a standardized, reusable resource for the community to study identity-preserving reconstruction in extreme conditions.  

InjuredFaces consists of $755$ injured facial portraits, denoted $I_{\text{inj}}$, corresponding to $N = 449$ distinct identities.
For each identity, we additionally provide a set of uninjured reference portraits, denoted $I_{\text{ref}}$, resulting in a total of $5{,}213$ reference images across the dataset.
These uninjured portraits serve as identity anchors and allow the construction of a stable reference representation for each individual.

For each injured image, we associate a formatted textual prompt $P$.
The prompt combines a fixed identity-neutral description, “a person’s face”, with a concise textual description of the visible injuries (e.g., “open wound on the forehead”).
This formulation enables controlled editing while keeping the identity description consistent across samples. 

To represent identity information, we rely on a pretrained face recognition embedding network.
Following prior work, we use \texttt{fal/AuraFace-v1} as our identity encoder, denoted $\mathcal{A}$.
Given an image $I$, the network maps it to a fixed-dimensional latent embedding $\mathcal{A}(I)$ that captures identity-related facial characteristics.
For each identity, we compute a reference embedding $a$ by averaging the embeddings of its corresponding uninjured portraits in $I_{\text{ref}}$.
These embeddings provide a consistent identity representation that we later use to assess identity preservation after reconstruction.
Representative examples from InjuredFaces are shown in Appendix~\ref{Anx:InjuredFaces}.

Facial reconstruction quality is assessed along five complementary dimensions: identity preservation, injury removal effectiveness, edit faithfulness, image quality, and perceptual similarity.

\textbf{Identity Preservation} is assessed using the identity embeddings defined above. Given a reconstructed image and its corresponding reference identity embedding, we measure preservation by computing the cosine similarity between their embeddings. This score reflects how well identity-related facial characteristics are retained after reconstruction. 
Examples of identity preservation values computed over representative image pairs, as well as the resulting score distributions, are reported in Appendix~\ref{Anx:InjuredFaces}.

\textbf{Injury Removal}
is assessed using a score derived from a large Vision–Language Model (VLM). Prior work has shown that large VLMs can be reliably repurposed for binary image classification tasks~\cite{kumari2025learning}. Given an edited image, we query the VLM with the question \emph{``Is this person's face injured? Only answer with yes or no.''} We then extract the logits corresponding to the \emph{Yes} and \emph{No} tokens and convert them into a scalar score using a sigmoid function: $\mathrm{VLMScore} = \sigma(l_{\text{Yes}} - l_{\text{No}})$.
Lower scores indicate more successful injury removal. This metric provides a robust, semantically grounded measure of injury presence.

\textbf{Edit Faithfulness} is measured using the CLIP score, computed as the cosine similarity between the CLIP embedding of the edited image and that of the target prompt. This measure captures how well the edited image aligns with the intended textual description.

\textbf{Image Quality} is evaluated in terms of visual realism and distributional consistency using FID \cite{heusel2017gans} and CMMD \cite{yang2024cmmd} (which has been shown to better correlate with human judgments of image quality). Both metrics are computed with respect to a reference set of 14{,}000 images from FFHQ. 

\textbf{Perceptual Similarity} is quantified using LPIPS \cite{zhang2018unreasonable} between the original and edited images. This metric complements identity preservation by capturing low-level visual deviations introduced by the editing process.  

To ensure a fair evaluation, IP, FID, CMMD, and LPIPS are computed only on samples where the edit is successful, defined as $\mathrm{VLMScore}<0.5$.
This prevents trivial preservation scores resulting from failed edits.
The proportion of successful edits is given in Appendix~\ref{experiment_details}.

\begin{figure*}[t!]
\centering
\setlength{\tabcolsep}{0pt}
\renewcommand{\arraystretch}{0.5}
\setlength{\gridimg}{0.14\textwidth}

\begin{tabular}{c @{\hspace{5pt}} c@{}c@{}c@{}c@{}c@{}c}
% Header row
\scriptsize{Input} & \scriptsize{FlowID} & \scriptsize{Kontext} & \scriptsize{RF-Solver} & \scriptsize{ICEdit} & \scriptsize{UltraEdit} & \scriptsize{SDEdit} \\[2pt]
% Alex Saucedo
\includegraphics[width=\gridimg]{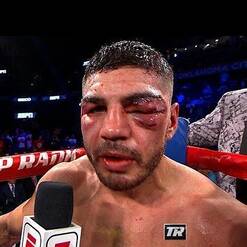} &
\includegraphics[width=\gridimg]{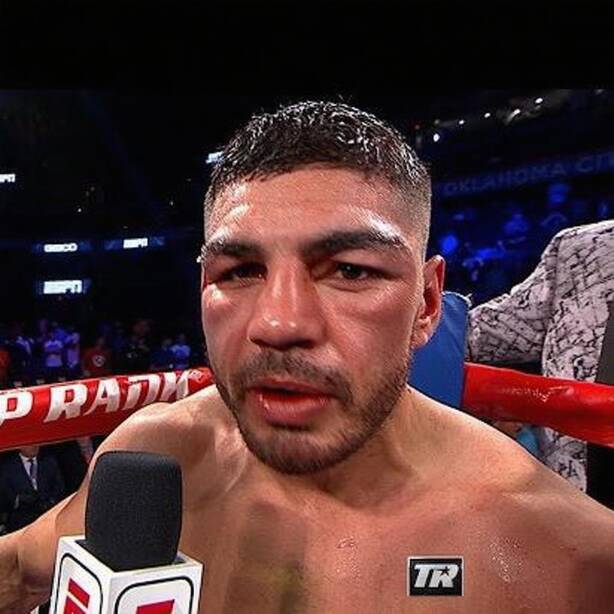} &
\includegraphics[width=\gridimg]{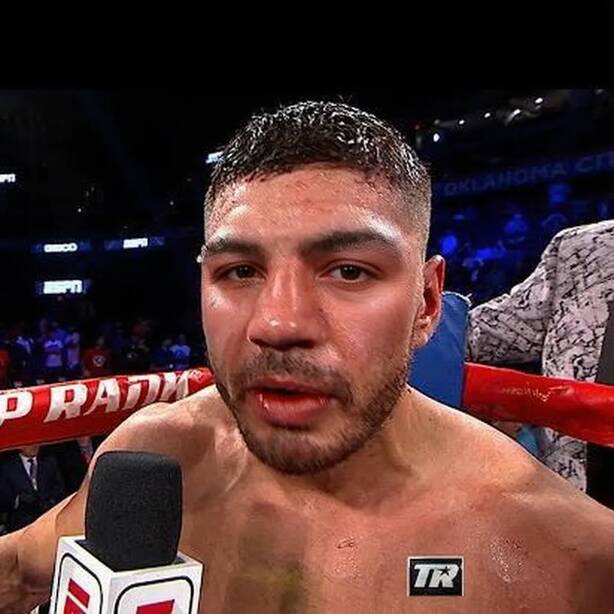} &
\includegraphics[width=\gridimg]{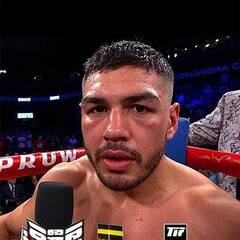} &
\includegraphics[width=\gridimg]{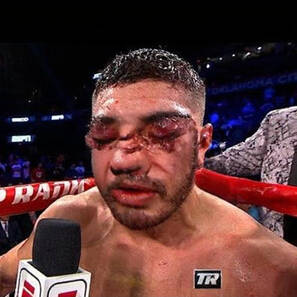} &
\includegraphics[width=\gridimg]{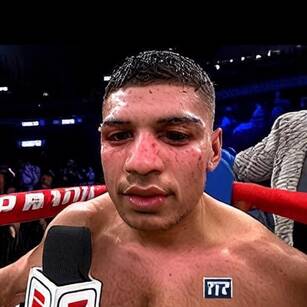} &
\includegraphics[width=\gridimg]{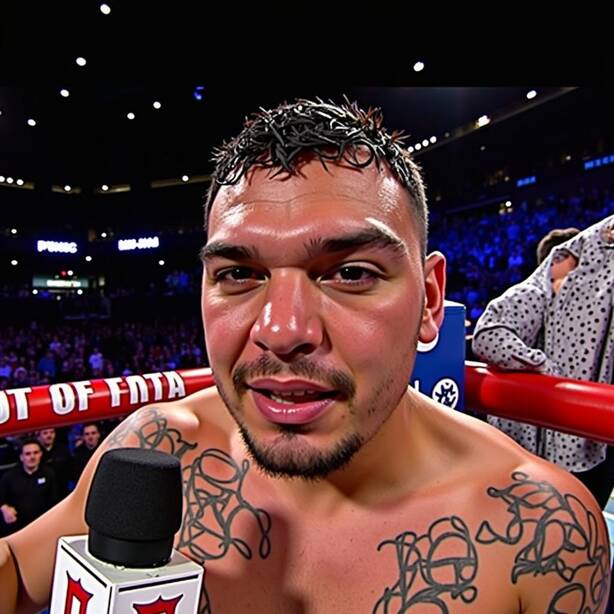} \\
% DEIVESON FIGUEIREDO
\includegraphics[width=\gridimg]{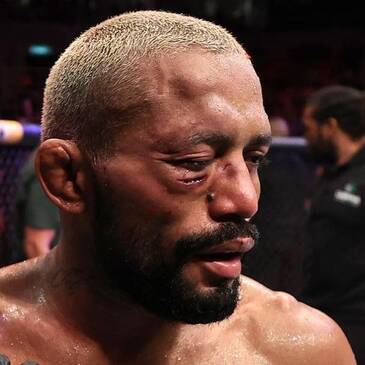} &
\includegraphics[width=\gridimg]{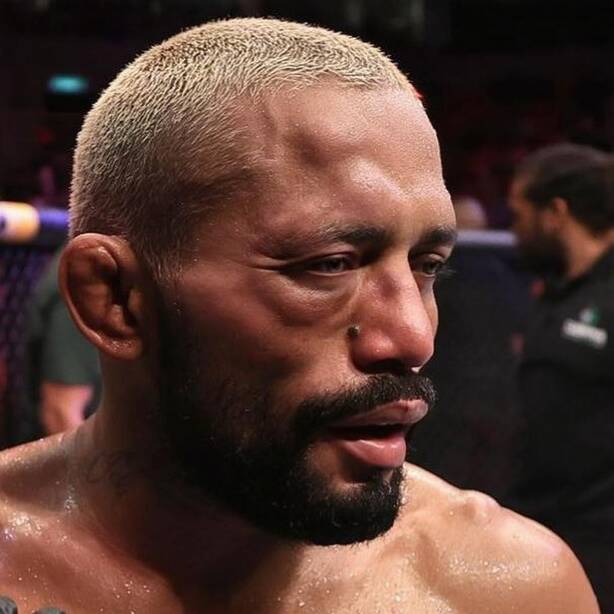} &
\includegraphics[width=\gridimg]{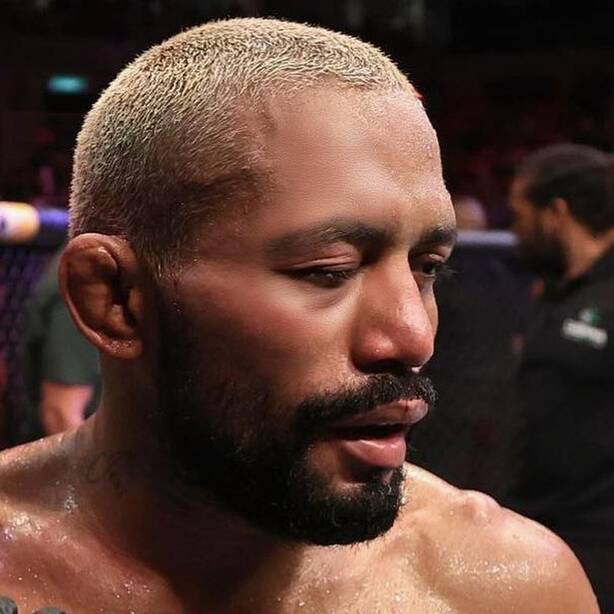} &
\includegraphics[width=\gridimg]{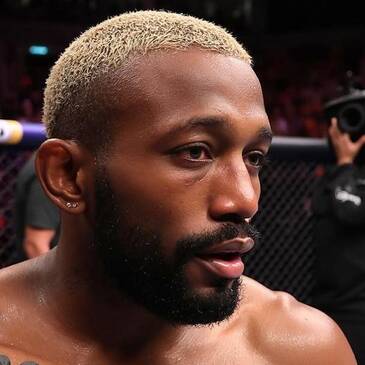} &
\includegraphics[width=\gridimg]{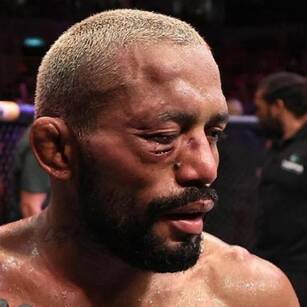} &
\includegraphics[width=\gridimg]{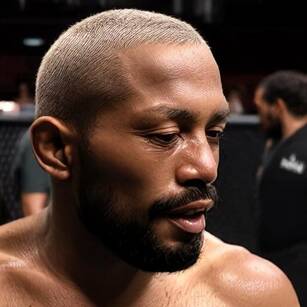} &
\includegraphics[width=\gridimg]{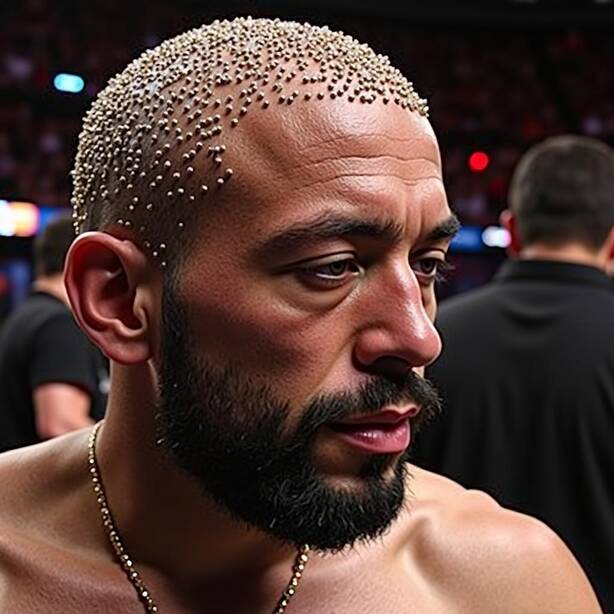} \\
% GIGA CHIKADZE
\includegraphics[width=\gridimg]{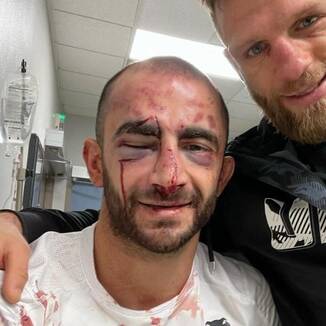} &
\includegraphics[width=\gridimg]{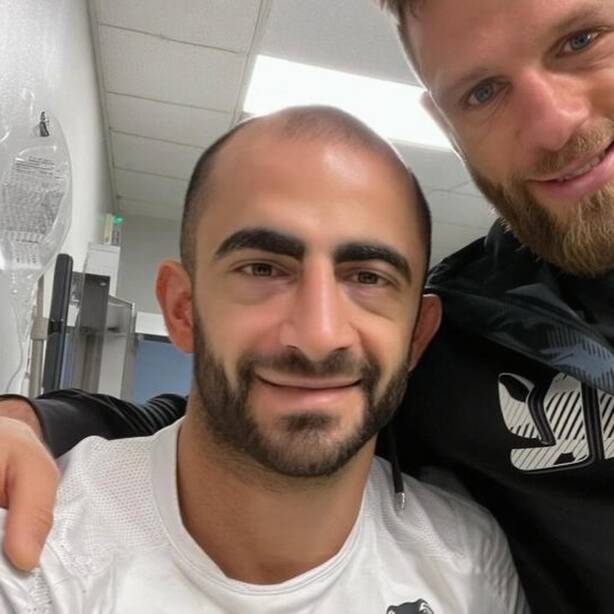} &
\includegraphics[width=\gridimg]{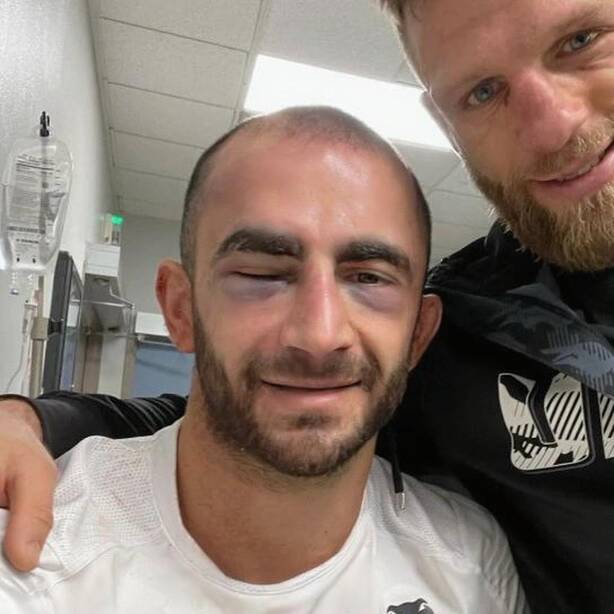} &
\includegraphics[width=\gridimg]{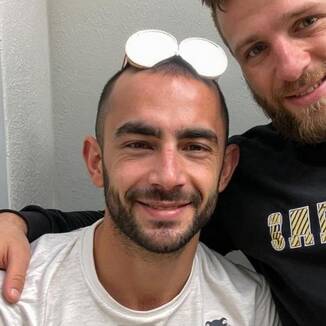} &
\includegraphics[width=\gridimg]{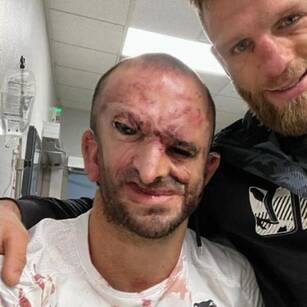} &
\includegraphics[width=\gridimg]{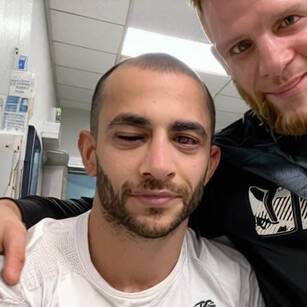} &
\includegraphics[width=\gridimg]{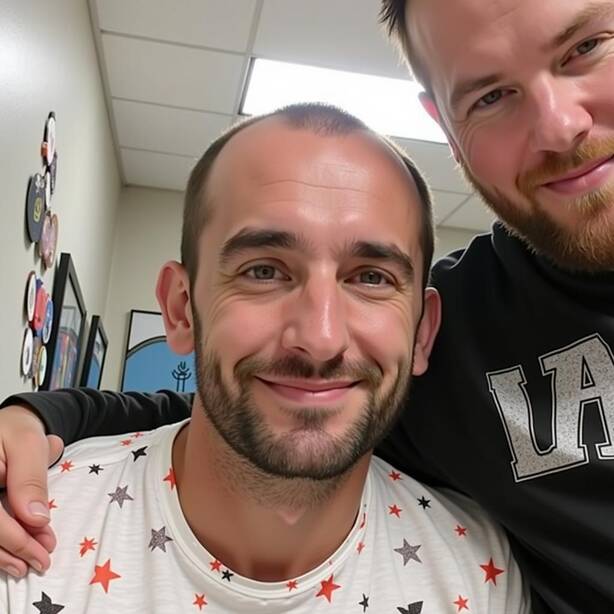} \\
% JIM MILLER
\includegraphics[width=\gridimg]{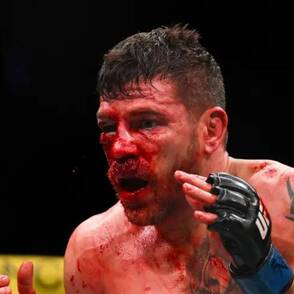} &
\includegraphics[width=\gridimg]{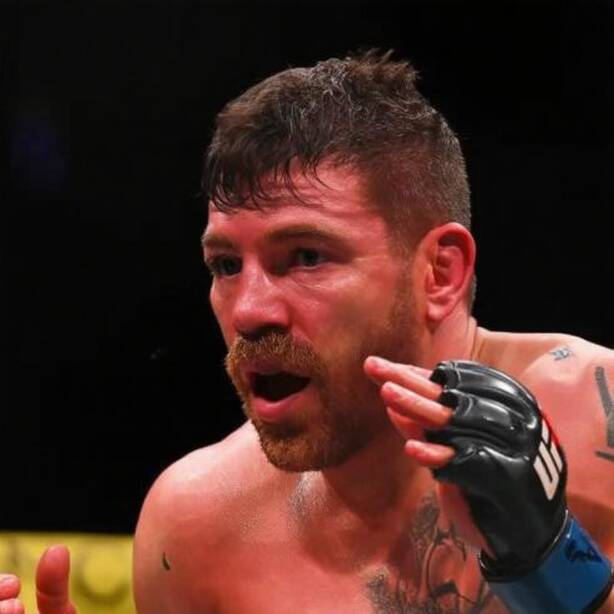} &
\includegraphics[width=\gridimg]{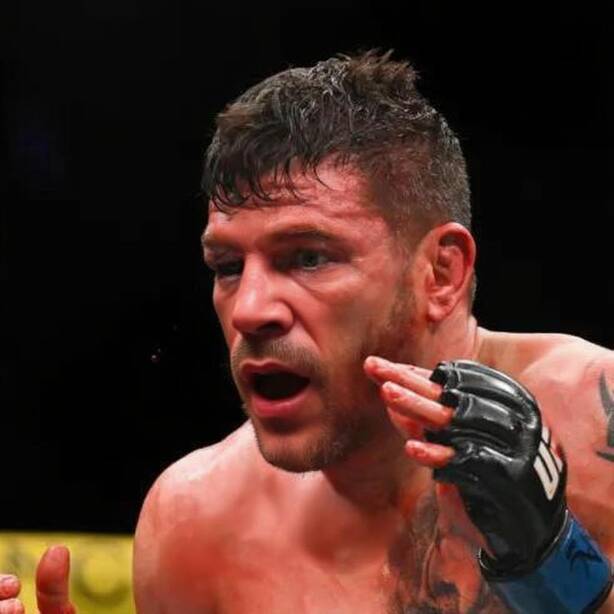} &
\includegraphics[width=\gridimg]{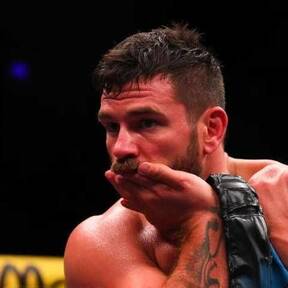} &
\includegraphics[width=\gridimg]{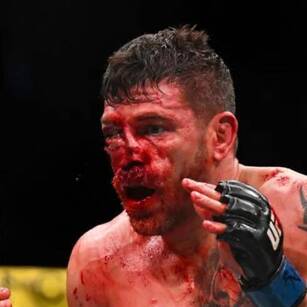} &
\includegraphics[width=\gridimg]{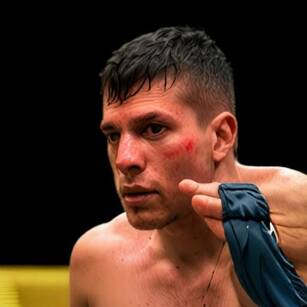} &
\includegraphics[width=\gridimg]{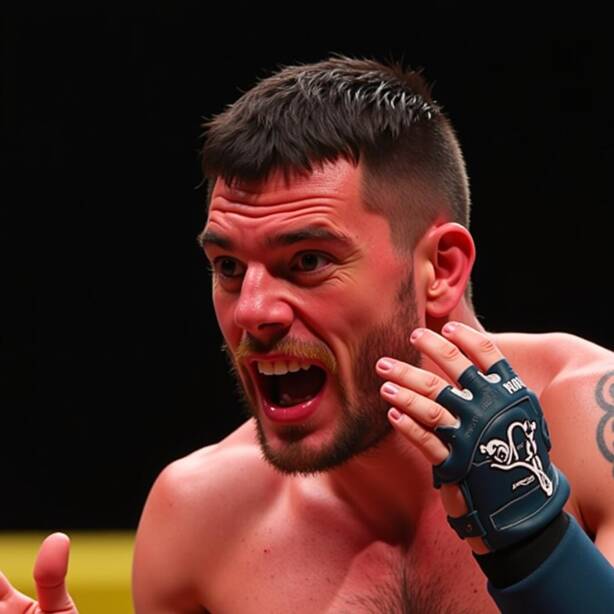} \\
% ASHLEE 
\includegraphics[width=\gridimg]{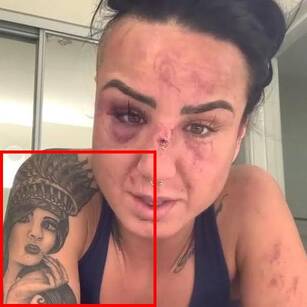} &
\includegraphics[width=\gridimg]{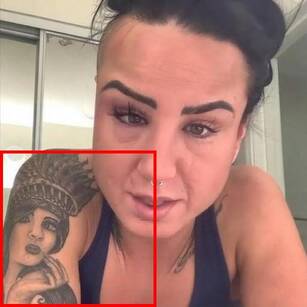} &
\includegraphics[width=\gridimg]{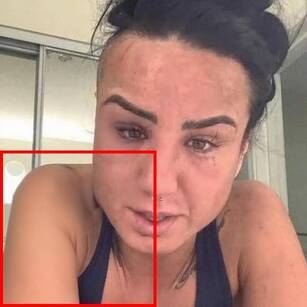} &
\includegraphics[width=\gridimg]{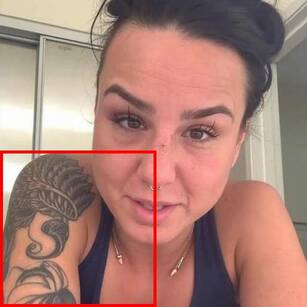} &
\includegraphics[width=\gridimg]{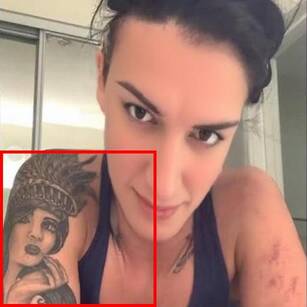} &
\includegraphics[width=\gridimg]{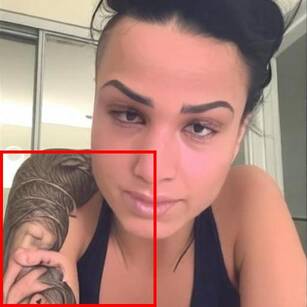} &
\includegraphics[width=\gridimg]{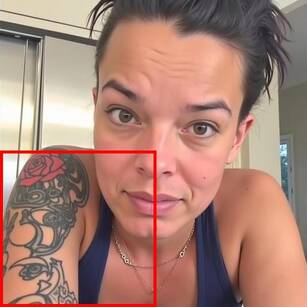} \\
% ASHLEE cropped
\includegraphics[width=\gridimg]{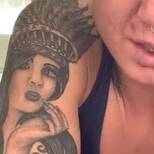} &
\includegraphics[width=\gridimg]{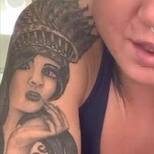} &
\includegraphics[width=\gridimg]{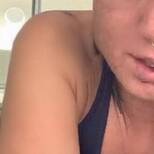} &
\includegraphics[width=\gridimg]{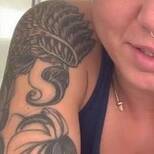} &
\includegraphics[width=\gridimg]{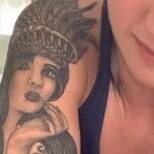} &
\includegraphics[width=\gridimg]{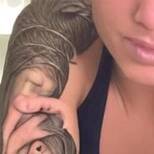} &
\includegraphics[width=\gridimg]{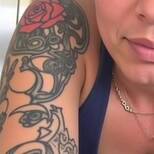} \\
\end{tabular}
\caption{Qualitative comparison of image editing methods on InjuredFaces.}
\label{fig:method_comparison_1}
\end{figure*}

\subsection{Results}

\textbf{Qualitatively}, Figure~\ref{fig:method_comparison_1} presents a comparison for different individuals from the InjuredFaces benchmark.
FlowID consistently achieves a better balance between injury removal and identity preservation.
It removes severe facial injuries while introducing minimal unintended changes to the original appearance.
Methods relying on heavier backbones, such as Kontext and RF-Solver, often succeed at removing visible injuries but struggle to preserve identity.
For instance, in the first and fourth rows, RF-Solver removes the targeted injury but alters global facial structure, leading to noticeable identity drift.
In addition, the tattoo close-up in the last row shows that several baselines inadvertently deform identity-relevant details, whereas FlowID better preserves such fine-grained cues.
% Other baselines fail for complementary reasons.
SDEdit applies strong edits but suffers from substantial identity degradation due to its non-inversion-based formulation.
UltraEdit and ICEdit, although effective on in-distribution editing tasks, perform poorly in this setting.
Both methods are trained on curated editing pairs and struggle with the out-of-distribution nature of severe facial trauma.
ICEdit, despite relying on a strong Flux-based backbone, remains constrained by its training regime and lacks robustness for unseen forensic edits.

\textbf{Quantitatively}, Table~\ref{tab:comparison}(a) reports results on the full InjuredFaces benchmark.
FlowID achieves the highest identity preservation score among all evaluated methods, demonstrating superior retention of identity-critical facial features after reconstruction.
This result is particularly significant given the severity of the edits required for injury removal.
At the same time, FlowID delivers markedly better injury removal performance.
It attains the lowest (best) VLM score and the best CLIP alignment by a wide margin, indicating both effective suppression of injury-related artifacts and strong faithfulness to the target prompt.
FlowID ranks second in terms of LPIPS, behind Kontext; this difference is consistent with its substantially higher injury removal success, as stronger and more complete edits necessarily induce larger perceptual changes.
Importantly, these improvements do not sacrifice identity fidelity, in contrast to competing approaches.
FlowID further maintains FID and CMMD on par with those of other methods.
Beyond metric improvements, these results are directly relevant to forensic identification.
Reliable suppression of injury-related artifacts, combined with strong identity preservation, increases the likelihood that reconstructed images remain recognizable by next-of-kin while reducing unnecessary visual distress.

\subsection{Additional Results on FFHQ}
To evaluate the capacity of FlowID to generalise beyond injury reconstruction, we report additional results on a subset of the FFHQ dataset focusing on classical facial attribute editing such as hair color, beard, glasses, and facial expression.
For each image, we apply a single textual edit instruction while keeping the original image as reference.
This setting allows us to assess whether FlowID maintains strong identity preservation and localized control in standard, non-forensic editing scenarios.
Table~\ref{tab:comparison}(b) reports results on single-shot edits.
In this setting, FlowID achieves the highest identity preservation score among all methods, while maintaining VLM edit score (higher is better) comparable to Kontext, indicating strong generalization beyond injury-specific edits.
In addition, FlowID attains the lowest FID, CMMD, and LPIPS values, indicating superior visual fidelity and minimal unintended perceptual changes under successful edits.

Qualitative examples and additional results on sequential editing can be found in our project page\footnote{\url{https://jrpll.github.io/flowid/}}.
\begin{table*}[ht]
\centering
\small
\setlength{\tabcolsep}{3pt}
\begin{minipage}{0.48\textwidth}
    \centering
    \begin{tabular}{@{}lcccccc@{}}
    \toprule
    Method & IP $\uparrow$ & CLIP $\downarrow$ & FID $\downarrow$ & CMMD $\downarrow$ & LPIPS $\downarrow$ & VLM $\downarrow$ \\
    \midrule
    UltraEdit & 0.30 & 0.23 & 116.44 & 2.36 & 0.20 & 0.19 \\
    SDEdit & 0.23 & 0.24 & \textbf{104.48} & \textbf{2.29} & 0.24 & 0.15 \\
    ICEdit & 0.22 & 0.25 & 190.43 & 2.61 & 0.24 & 0.88 \\
    Kontext & 0.48 & 0.24 & 128.78 & 3.47 & \textbf{0.10} & 0.28 \\
    RF-Solver & 0.48 & 0.24 & 117.88 & 3.12 & 0.14 & 0.36 \\
    FlowID & \textbf{0.50} & \textbf{0.23} & 123.07 & 3.09 & 0.12 & \textbf{0.14} \\ %0.46 lpips
    \bottomrule
    \end{tabular}
    \par\vspace{4pt}
    \scriptsize (a) InjuredFaces
\end{minipage}%
\hfill
\begin{minipage}{0.48\textwidth}
    \centering
    \begin{tabular}{@{}lcccccc@{}}
    \toprule
    Method & IP $\uparrow$ & CLIP $\uparrow$ & FID $\downarrow$ & CMMD $\downarrow$ & LPIPS $\downarrow$ & VLM $\uparrow$ \\
    \midrule
    UltraEdit & 0.46 & 0.26 & 35.28 & 0.83 & 0.24 & 0.84 \\
    SDEdit & 0.13 & 0.28 & 69.10 & 1.80 & 0.49 & 0.61 \\
    ICEdit & 0.75 & 0.27 & 36.37 & 0.58 & 0.19 & 0.65 \\
    Kontext & 0.72 & 0.27 & 31.30 & 0.85 & 0.20 & \textbf{0.96} \\
    RF-Solver & 0.26 & \textbf{0.28} & 51.59 & 1.38 & 0.31 & 0.87 \\
    FlowID & \textbf{0.77} & 0.27 & \textbf{29.04} & \textbf{0.45} & \textbf{0.11} & 0.94 \\
    \bottomrule
    \end{tabular}
    \par\vspace{4pt}
    \scriptsize (b) FFHQ
\end{minipage}
\caption{Quantitative comparison on (a) InjuredFaces and (b) FFHQ.}
\label{tab:comparison}
\end{table*}

\begin{table}[tb]
\centering
\small
\setlength{\tabcolsep}{3pt}
\begin{tabular}{@{}lcccccc@{}}
\toprule
Method & IP $\uparrow$ & CLIP $\uparrow$ & FID $\downarrow$ & CMMD $\downarrow$ & LPIPS $\downarrow$ & VLM $\uparrow$ \\
\midrule
w/o fine-tuning & 0.52 & 0.27 & 32.32 & 1.08 & 0.15 & \textbf{0.99} \\
w/o masking & 0.61 & 0.27 & 34.95 & 0.63 & 0.24 & 0.96 \\
FlowID & \textbf{0.77} & 0.27 & \textbf{29.04} & \textbf{0.45} & \textbf{0.11} & 0.94 \\
\bottomrule
\end{tabular}
\caption{Ablation on fine-tuning and masking components.}
\label{tab:ablation}
\end{table}

\subsection{Ablation Study}

\begin{figure}[t]
\centering
\setlength{\tabcolsep}{1pt}
\setlength{\imgwidth}{0.24\columnwidth}
\begin{tabular}{@{}cccc@{}}
\begin{subfigure}[b]{\imgwidth}
    \centering
    \caption*{\scriptsize }
    \includegraphics[width=\textwidth]{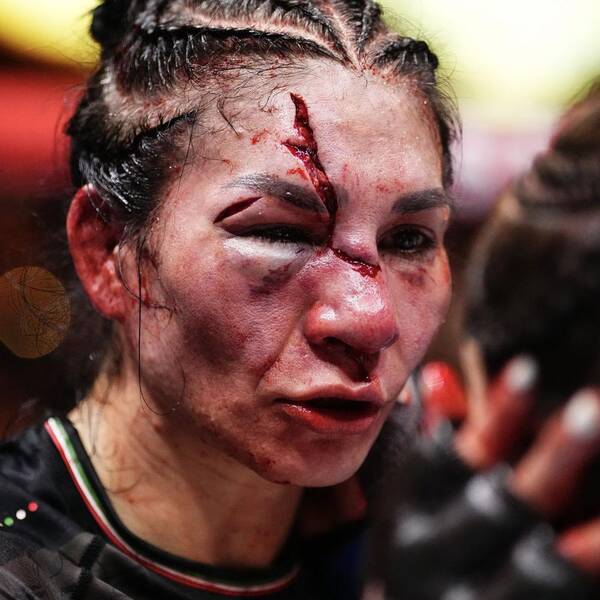}
    \caption*{\scriptsize Input}
\end{subfigure} &
\begin{subfigure}[b]{\imgwidth}
    \centering
    \caption*{\scriptsize $\mathrm{IP}=0.27$}
    \includegraphics[width=\textwidth]{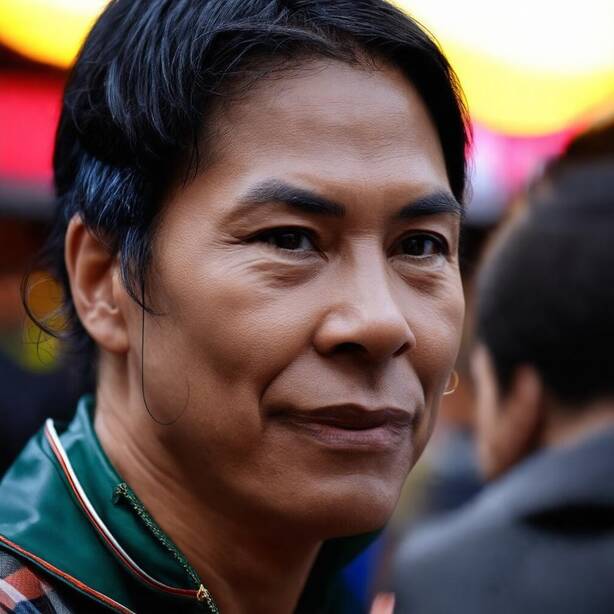}
    \caption*{\scriptsize $N{=}0$}
\end{subfigure} &
\begin{subfigure}[b]{\imgwidth}
    \centering
    \caption*{\scriptsize 0.42}
    \includegraphics[width=\textwidth]{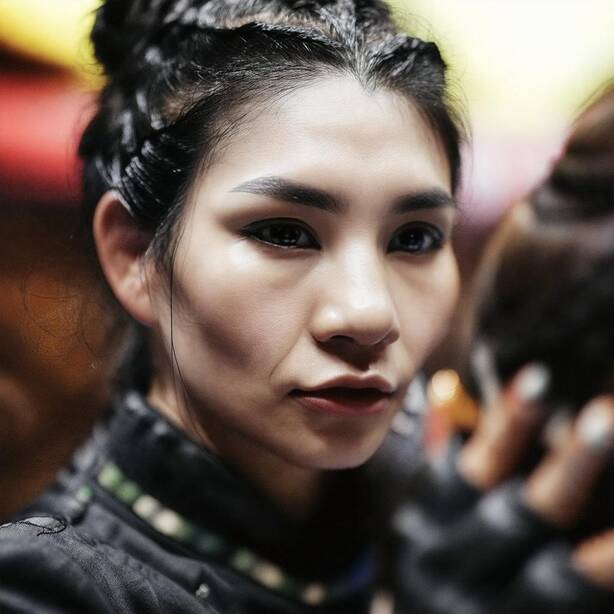}
    \caption*{\scriptsize 50}
\end{subfigure} &
\begin{subfigure}[b]{\imgwidth}
    \centering
    \caption*{\scriptsize 0.56}
    \includegraphics[width=\textwidth]{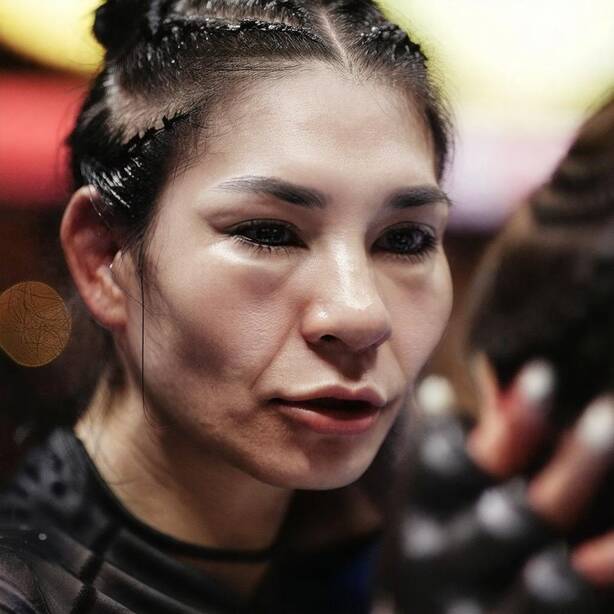}
    \caption*{\scriptsize 100}
\end{subfigure}
\end{tabular}
\caption{Edit performance as a function of fine-tuning steps.}
\label{fig:ft_impact}
\end{figure}

To assess the contribution of each component, we perform an ablation study on FFHQ by removing fine-tuning and masking individually. Results are reported in Table~\ref{tab:ablation}.

Without fine-tuning, identity preservation drops substantially (0.52 vs.\ 0.77), and CMMD increases (1.08 vs.\ 0.45), indicating weaker semantic and distributional alignment. Nevertheless, even without fine-tuning, the method already matches or exceeds several competing approaches in identity preservation on FFHQ (e.g., outperforming UltraEdit at 0.46 and SDEdit at 0.13).

The effect of single-image fine-tuning is further illustrated in Figure~\ref{fig:ft_impact}. As the number of fine-tuning steps increases, identity preservation improves steadily, rising from 0.27 without fine-tuning to 0.56 after 100 steps. Qualitatively, fine-tuning progressively restores identity-defining facial traits such as facial structure and proportions, while preserving the effectiveness of injury removal.

Removing masking degrades realism and perceptual quality, with FID increasing from 29.04 to 34.95 and LPIPS from 0.11 to 0.24, confirming that unconstrained edits introduce unnecessary visual distortions. Despite this degradation, the unmasked variant still achieves stronger identity preservation (0.61) than most baselines, including UltraEdit and SDEdit, while maintaining a high editing success rate (VLM 0.96).

Figure~\ref{fig:mask-based-preservation} illustrates the effect of localized masking. Without masking, the model alters regions unrelated to the injury, degrading fine-grained identity cues such as tattoos and increasing perceptual distortion, whereas applying the inferred mask enforces exact reconstruction outside the editable region, leading to improved LPIPS and FID while preserving identity-critical details.

When combined, fine-tuning and masking yield consistent improvements across all metrics, achieving the best trade-off between identity preservation, image quality, and edit success. The slight decrease in success rate with masking reflects a deliberate trade-off: restricting edits to semantically relevant regions limits prompt flexibility but results in  higher fidelity and perceptual quality. 
\section{Conclusion}

In many medico-legal and humanitarian identification workflows, facial imagery is often the only usable visual evidence. Severe trauma, post-mortem degradation, and occlusions can remove key facial cues, slowing identification, increasing forensic workload, and prolonging uncertainty for families. This paper re-frames facial reconstruction as a recognition focused task that prioritizes identity preservation and strict edit containment, and makes two primary contributions.

First, we introduce FlowID, an identity preserving facial editing framework that restores recognizable cues while constraining unintended changes. FlowID combines per-image fine-tuning, adapting the generator to each image, with an attention-derived masking mechanism that localizes edits to damaged regions while enforcing exact reconstruction elsewhere. Second, we propose InjuredFaces, a benchmark and evaluation protocol for reconstruction under severe facial trauma, where identity preservation is the primary objective.

FlowID is validated experimentally and has already shown operational impact. It has contributed to successful identifications, including those of deceased migrants recovered in the Mediterranean, and has been deployed across several medico-legal sites, in Mexico and Colombia. These deployments highlight the practical relevance of identity centric reconstruction in reducing processing time and cognitive burden in resource constrained institutions, and may also help limit repeated exposure to graphic imagery.

To support reproducible progress, InjuredFaces provides a standardized benchmark aligned with realistic forensic conditions. Unlike evaluations focused on generic realism or stylistic fidelity, it centers around identity preservation, injury removal, and unintended modifications, encouraging methods better matched to high stakes identification workflows.

FlowID has limitations. Per-image fine-tuning adds computational overhead, though it remains lightweight, can be reduced through parameter efficient adaptation such as LoRA style updates, and amortizes over sequential edits since later modifications run at standard generation cost. Furthermore, our masking component can prove less useful when injuries are globally present on the face. Improved masking integration and lower cost adaptation are our future work.

\appendix

\begin{table}[tb]
\centering
\small
\setlength{\tabcolsep}{3pt}
\begin{tabular}{@{}lccccccc@{}}
\toprule
& \multicolumn{5}{c}{FFHQ} & & InjuredFaces \\
\cmidrule(lr){2-6} \cmidrule(lr){8-8}
Method & Step 1 & Step 2 & Step 3 & Step 4 & Step 5 & & Step 1 \\
\midrule
UltraEdit & 0.85 & 0.84 & 0.83 & 0.80 & 0.78 & & 0.84 \\
SDEdit    & 0.58 & 0.58 & 0.57 & 0.56 & 0.56 & & 0.86 \\
ICEdit    & 0.64 & 0.58 & 0.54 & 0.46 & 0.43 & & 0.10 \\
Kontext   & 0.97 & 0.96 & \textbf{0.96} & 0.93 & 0.92 & & 0.52 \\
RF-Solver  & 0.88 & 0.86 & 0.86 & 0.85 & 0.85 & & 0.67 \\
FlowID    & \textbf{0.97} & \textbf{0.96} & 0.95 & \textbf{0.94} & \textbf{0.93} & & \textbf{0.95} \\
\bottomrule
\end{tabular}
\caption{Successful edit rates per method and benchmark.}
\label{tab:success_rates}
\end{table}

\section{Implementation Details}
\label{experiment_details}
Both InjuredFaces and FFHQ benchmarks were run on a A100 GPU cluster.
For methods that use source and target prompts like ours, we use the prompt describing the injured photo from InjuredFaces as the source, and \emph{"a person's face"} as the target.
For instruction-based methods, we infer the edit instructions based on the source and target prompt with simple rules.
For implementations that allow negative guidance, we extract the injury description from the source prompt by removing \emph{"a person's face"} from the prompt.

Our method uses Adam8Bit optimizer in order to lower the memory requirements for compatibility with consumer hardware.
We run all fine-tunings with batch size of 1, gradient accumulation of 10, 80 optimizer steps and use a learning rate of 5e-5.
These details allow us to run FlowID on a NVIDIA RTX4090 GPU.

For instruction-based methods we check for the presence of a given type of injury and turn it into an instruction, \textit{e.g}, if ``wound'' is in the description, the instruction becomes ``remove the wound''. Checked attributes are : ``wound'', ``blood'', ``mouth open'', ``bump'', ``bruise'', `cut`'', ``injury'', ``swollen''.
For ICEdit, we used the following prompt : ``A diptych with two side-by-side images of the same scene. On the right, the scene is exactly the same as on the left but {instruction}'' where the instruction with the above rule.
For RF-Solver, we use default parameters except for the number of features injection steps, which we set to 10, as it yielded the best balance between identity preservation and edit insertion.
Similarly we used a strength $t_s$ of 0.65 for SDEdit.

Success rates for both sequential edits on FFHQ and one-step edits on InjuredFaces can be seen in Table~\ref{tab:success_rates}. We notice that our method tends to edit more strongly than others, thus motivating the need for the filtering mentioned in Section~\ref{sec:injuredfaces}.

\section{Additional Details on InjuredFaces}
\label{Anx:InjuredFaces}
\begin{figure}[t]
\centering
\scalebox{1.4}{% <-- Change this value: 1.5 = 150%, 0.8 = 80%, etc.
\begin{tikzpicture}
    \def\imgsize{1cm}
    \def\labely{-0.7}
    
    % injured_0.jpg
    \node[inner sep=0] (img0) at (0,0) {
        \includegraphics[width=\imgsize, height=\imgsize]{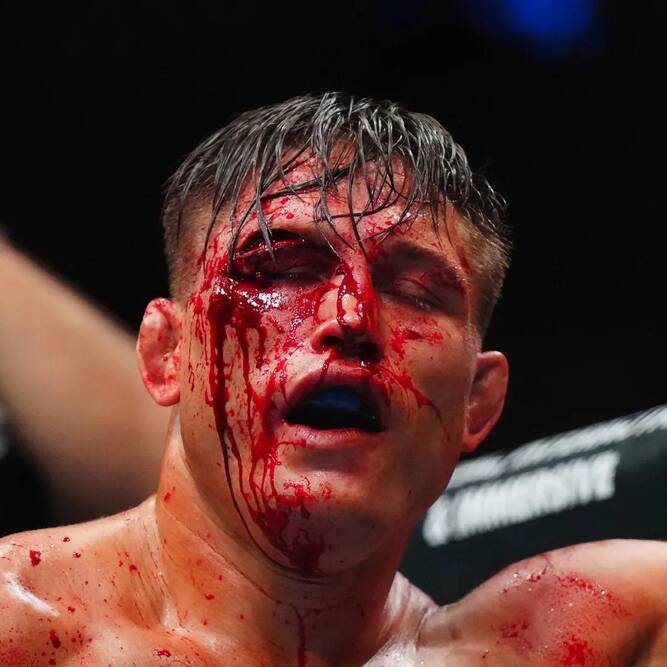}
    };
    \node[font=\scriptsize] at (0,\labely) {$I_{inj}$};
    
    % 0.jpg
    \node[inner sep=0] (img1) at (1.1,0) {
        \includegraphics[width=\imgsize, height=\imgsize]{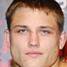}
    };
    
    % dots
    \node (dots) at (2.0,0) {$\cdots$};
    \node[font=\scriptsize] at (2.0,\labely) {$I_{ref}$};
    
    % 3.jpg
    \node[inner sep=0] (img2) at (2.9,0) {
        \includegraphics[width=\imgsize, height=\imgsize]{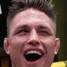}
    };
    
    % caption text
    \node[font=\scriptsize, text width=1.3cm, align=center] (caption) at (4.1,0) {\emph{"a person's face, injured"}};
    \node[font=\scriptsize] at (4.1,\labely) {$P$};
    
    % embedding tensor
    \draw[fill=blue!5, line width=0.4pt] (5.2,-0.5) rectangle (5.45,-0.3);
    \draw[fill=blue!5, line width=0.4pt] (5.2,-0.3) rectangle (5.45,-0.1);
    \draw[fill=blue!5, line width=0.4pt] (5.2,-0.1) rectangle (5.45,0.1);
    \draw[fill=blue!5, line width=0.4pt] (5.2,0.1) rectangle (5.45,0.3);
    \draw[fill=blue!5, line width=0.4pt] (5.2,0.3) rectangle (5.45,0.5);
    \node[font=\scriptsize] at (5.325,\labely) {$a$};
    
\end{tikzpicture}
}% end scalebox
\caption{An InjuredFaces sample.}
\label{fig:sample_data}
\end{figure}

Figure \ref{fig:sample_data} shows a sample from our InjuredFaces dataset.
Each injured faces represents a row in our dataset.
It comes with healthy portraits of the same person, a textual description of the injured portraits, and an embedding vector $a$, computed as the average of identity embeddings. 

\begin{figure}[t]
\centering
\resizebox{\columnwidth}{!}{%
\begin{subfigure}[b]{\columnwidth}
    \centering
    \includegraphics[width=0.8\linewidth]{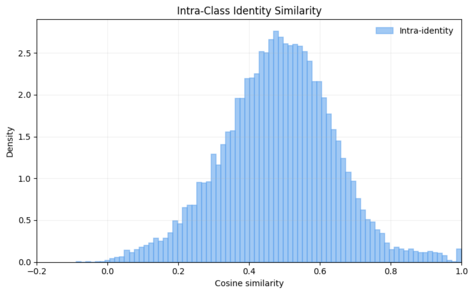}
    \caption{Intra-class identity similarity distribution.}
    \label{fig:similarity_distributions}
\end{subfigure}%
}

\vspace{1em}

\resizebox{\columnwidth}{!}{%
\begin{subfigure}[b]{\columnwidth}
    \centering
    \begin{tikzpicture}
        \node[inner sep=0] (matrix) {\includegraphics[width=5.5cm]{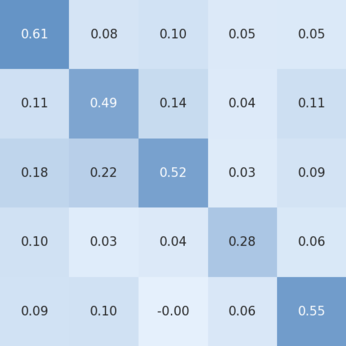}};
        
        % Column images (loop unrolled so arXiv's dependency scanner detects the files)
        \node[inner sep=0, above=2pt of matrix.north, anchor=south, xshift={-2.2cm + 0*1.1cm}]
            {\includegraphics[width=1.1cm]{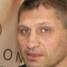}};
        \node[inner sep=0, above=2pt of matrix.north, anchor=south, xshift={-2.2cm + 1*1.1cm}]
            {\includegraphics[width=1.1cm]{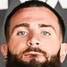}};
        \node[inner sep=0, above=2pt of matrix.north, anchor=south, xshift={-2.2cm + 2*1.1cm}]
            {\includegraphics[width=1.1cm]{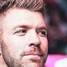}};
        \node[inner sep=0, above=2pt of matrix.north, anchor=south, xshift={-2.2cm + 3*1.1cm}]
            {\includegraphics[width=1.1cm]{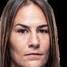}};
        \node[inner sep=0, above=2pt of matrix.north, anchor=south, xshift={-2.2cm + 4*1.1cm}]
            {\includegraphics[width=1.1cm]{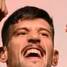}};

        % Row images (loop unrolled so arXiv's dependency scanner detects the files)
        \node[inner sep=0, left=2pt of matrix.west, anchor=east, yshift={2.2cm - 0*1.1cm}]
            {\includegraphics[width=1.1cm]{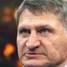}};
        \node[inner sep=0, left=2pt of matrix.west, anchor=east, yshift={2.2cm - 1*1.1cm}]
            {\includegraphics[width=1.1cm]{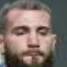}};
        \node[inner sep=0, left=2pt of matrix.west, anchor=east, yshift={2.2cm - 2*1.1cm}]
            {\includegraphics[width=1.1cm]{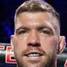}};
        \node[inner sep=0, left=2pt of matrix.west, anchor=east, yshift={2.2cm - 3*1.1cm}]
            {\includegraphics[width=1.1cm]{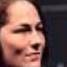}};
        \node[inner sep=0, left=2pt of matrix.west, anchor=east, yshift={2.2cm - 4*1.1cm}]
            {\includegraphics[width=1.1cm]{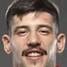}};
    \end{tikzpicture}
    \caption{Cosine similarity matrix between identity embeddings.}
    \label{fig:cosine_matrix}
\end{subfigure}%
}
\caption{Identity similarity analysis.}
\label{fig:identity_analysis}
\end{figure}

The histogram shown in Figure~\ref{fig:similarity_distributions} reports the intra-class identity similarity distribution computed over reference images, showing a stable and well-shaped distribution centered around moderate to high similarity values.

Figure~\ref{fig:cosine_matrix} provides additional insight into the identity preservation metric. The cosine similarity matrix illustrates pairwise identity similarities across multiple identities, with higher values along same-identity pairs and substantially lower values for mismatched identities. 

Together, these visualizations confirm that the identity embedding produces consistent representations for the same individual, supporting its use as a reference for identity preservation evaluation.

\section*{Ethical Statement}

This work raises important questions regarding the processing of images of deceased individuals without their consent. We acknowledge this tension, but note that it presents an inherent dilemma: the very purpose of altering these images is to identify the deceased and return them to their families. In the absence of other clues, involving communities through carefully processed images may be the only viable path toward recognition.
To ensure responsible deployment, all processing is conducted locally by authorities responsible for identification, and reconstructed images are shared only through official channels, clearly marked as renderings. We also recognize that reconstructions may mislead family members; our system is therefore designed to support—not replace—existing identification workflows, with final determinations always involving verification by forensic professionals.

\section*{Acknowledgements}
This work was supported by a French government grant managed by the Agence Nationale de la Recherche under the "Investissements d'avenir" program (reference "ANR-21-ESRE-0051").

\bibliographystyle{named}
\bibliography{ijcai26}

\end{document}